%% file: AAAI.tex
\def\year{2019}\relax
\documentclass[letterpaper]{article} 
\usepackage{aaai19}  
\usepackage{times}  
\usepackage{helvet}  
\usepackage{courier}  
\usepackage{url}  
\usepackage{graphicx}  

\usepackage{times}
\usepackage{url}
\usepackage{latexsym}
\usepackage{tabularx}
\usepackage{multirow}
\usepackage{diagbox}
\usepackage{graphicx}
\usepackage{extarrows}
\usepackage{algorithm}
\usepackage{algorithmicx}
\usepackage{amsmath}
\usepackage{amsmath,bm}
\usepackage{booktabs}
\usepackage{import}
\usepackage{enumerate}
\usepackage{float}
\usepackage{mathrsfs}
\usepackage{amsfonts}
\usepackage{caption}
\usepackage{xspace}
\usepackage{paralist}
\usepackage{color}
\usepackage{algorithmicx}
\usepackage{algpseudocode}
\usepackage{amsmath}
\usepackage{amsfonts}
\usepackage{arydshln}
\begin{document} 

\title{A Neural Multi-Task Learning Framework to Jointly Model \\Medical Named Entity Recognition and Normalization}


\author{
	Sendong Zhao\footnotemark[1], Ting Liu\footnotemark[3], Sicheng Zhao\footnotemark[2], Fei Wang\footnotemark[1]\\ 
	\footnotemark[1]~Weill Cornell Medical College, Cornell University, USA\\
	\footnotemark[2]~Department of Electrical Engineering and Computer Sciences, University of California Berkeley, USA\\
	\footnotemark[3]~Research Center for Social Computing and Information Retrieval, Harbin Institute of Technology, China  \\
	\{sez4001,few2001\}@med.cornell.edu, schzhao@gmail.com, tliu@ir.hit.edu.cn
}


\maketitle
\begin{abstract}
\input{0abstract}
\end{abstract}

\section{Introduction}
\label{sec:intro}
\input{1introduction}

\section{Related Work}
\label{sec:related}
\input{2related}

\section{Problem Definition}
\label{sec:definition}
\input{definition}

\section{Our Method}
\label{sec:method}
\input{3method}

\section{Experiments}
\label{sec:experiment}
\input{4experiment}

\section{Conclusion}
\label{sec:conclusion}
\input{5conclusion}

\section{Acknowledgments}
\label{sec:acknoledgments}
\input{6acknoledgments}

\bibliography{refer}
\bibliographystyle{aaai}

\end{document}

%% file: 0abstract.tex

State-of-the-art studies have demonstrated the superiority of joint modeling over pipeline implementation for medical named entity recognition and normalization due to the mutual benefits between the two processes. 
To exploit these benefits in a more sophisticated way, we propose a novel deep neural multi-task learning framework with explicit feedback strategies to jointly model recognition and normalization.
On one hand, our method benefits from the general representations of both tasks provided by multi-task learning.
On the other hand, our method successfully converts hierarchical tasks into a parallel multi-task setting while maintaining the mutual supports between tasks. 
Both of these aspects improve the model performance. 
Experimental results demonstrate that our method performs significantly better than state-of-the-art approaches on two publicly available medical literature datasets.

%% file: 1introduction.tex
Due to the large amount of electronically-available medical publications stored in databases such as PubMed, there has been an increasing interest in applying text mining and information extraction to the medical literature. 
Those techniques can generate tremendous benefits for both medical research and applications. Among the medical literature mining tasks, medical named entity recognition and normalization are the most fundamental tasks.

The goal of medical named entity recognition and normalization is to find the boundaries of mentioning from the medical text and map them onto a controlled vocabulary. State-of-the-art studies have demonstrated the superiority of joint modeling of medical named entity recognition and normalization compared to the pipeline implementation due to mutual benefits between them. There are two main limitations of pipeline models: (1) errors from the recognition tagging cascade into normalization errors, and (2) recognition and normalization are mutually useful to each other, but pipeline models cannot utilize these potential benefits. Joint modeling recognition and normalization can naturally alleviate these 
limitations and achieve better performance. For example, \citeauthor{Leaman2016TaggerOne} (\citeyear{Leaman2016TaggerOne}) leveraged a joint scoring function for medical named entity recognition and normalization. \citeauthor{Lou2017A} (\citeyear{Lou2017A}) proposed a transition-based model to jointly perform medical named entity recognition and normalization, casting the output construction process into an incremental state transition process. However, these existing joint modeling methods (1) rely heavily on hand-crafted features and task specific resources thus fail to encode complicated and general features such as character-level and semantic-level features; 
(2) use simplistic ways to jointly model medical named entity recognition and normalization, which cannot provide essential mutual supports between these two.   

\begin{figure}[tp]
	\centering
	\includegraphics[width=0.4\textwidth]{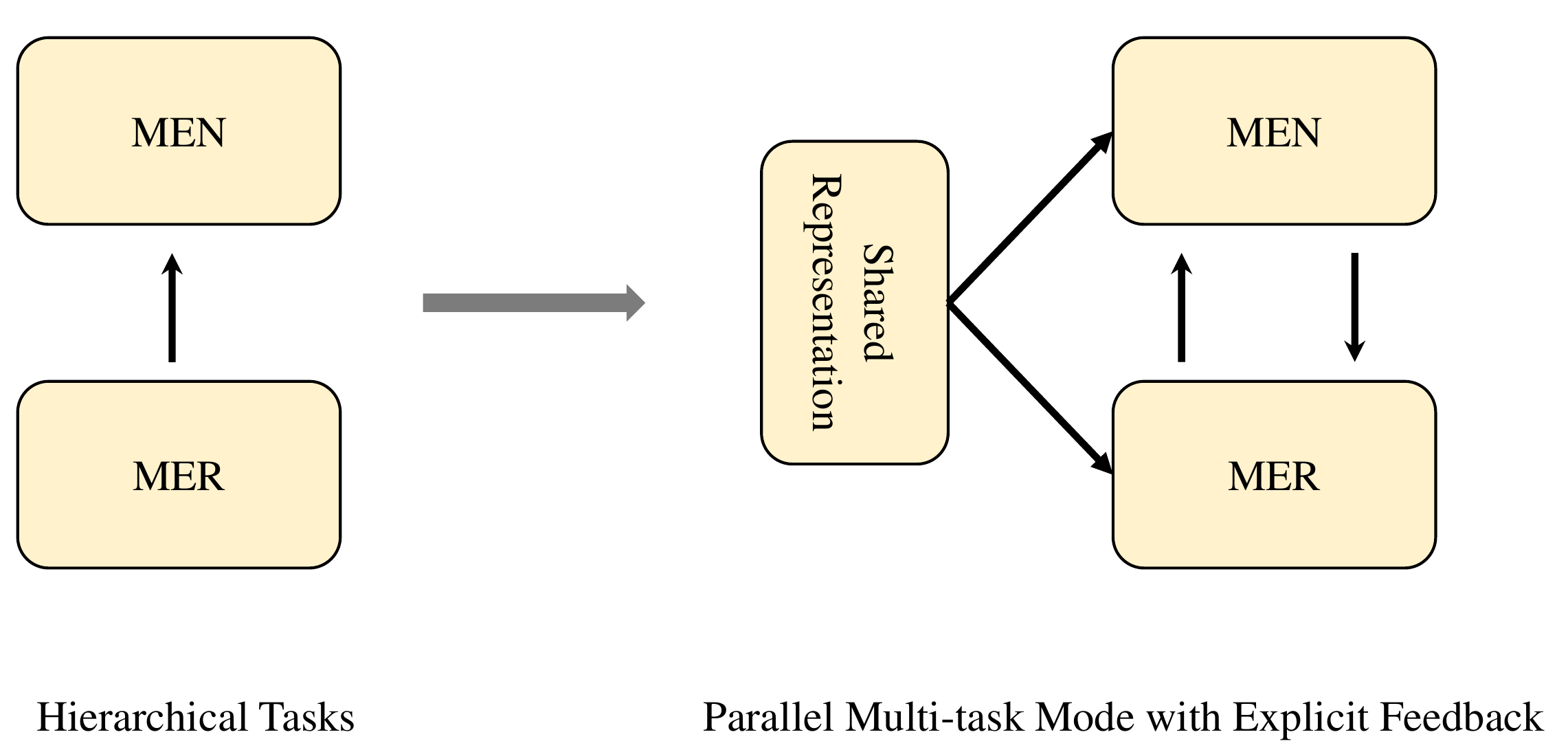}
	\vspace{-0.1in}
	\caption{From hierarchical tasks to parallel multi-task mode by incorporating explicit feedback strategies among tasks. 
		}\label{fig: trans}
	\vspace{-0.15in}
\end{figure}

To improve the joint modeling medical named entity recognition and normalization (MER and MEN), we propose a novel deep neural multi-task learning (MTL) framework with two explicit feedback strategies, which can make use of the mutual benefits between recognition and normalization in a more advanced and intelligent way. First, our method benefits from general representations of both tasks provided by multi-task learning, which enjoys a regularization effect~\cite{Collobert2011,DBLP:journals/corr/Ruder17a} that leads to more general representations to help both tasks. Specifically, it minimizes over-fitting to any specific tasks, thus makes the learned representations universal across tasks. Second, our method can successfully convert hierarchical tasks into a parallel multi-task setting while maintaining mutual supports between tasks. Although the general concept of deep neural multi-task learning is not new, the innovation of our method is that it incorporates both the feedback strategies from the low-level task to the high-level task and vice versa, as shown in Figure~\ref{fig: trans}. 
These two feedback strategies exploit the output of entity recognition to improve entity normalization and vice versa. In addition, our method uses Bi-LSTM to boost the sequential modeling of text and CNN to encode clues hidden in character-level features such as \textbf{Zo}lmitriptan, \textbf{Zomig} and \textbf{Zomig}on.

We evaluate our models across two corpora (the BioCreative V Chemical Disease Relation (BC5CDR) task corpus \cite{Li2016BioCreative} and the NCBI Disease corpus \cite{Rezarta2014NCBI}) of medical articles and outperform the state-of-the-art study by up to 4.53\% F1 on medical named entity recognition and 5.61\% F1 on medical named entity normalization.

\noindent\textbf{Contribution.}
To make use of the mutual benefits in a more sophisticated way, we propose a novel deep neural multi-task learning framework with explicit feedback strategies to jointly model named medical entity recognition and normalization. This method incorporates both the feedback strategies from the low-level task to the high-level task and vice versa, which makes it possible to convert hierarchical tasks, i.e. MER and MEN, into parallel multi-task mode while maintaining mutual supports between tasks.

%% file: 2related.tex
\textbf{MER and MEN.} 
Several existing studies typically run a medical named entity recognition model to extract entity names first, then run a medical named entity normalization model to link extracted names to a controlled vocabulary~\cite{Doan2010,Sahu2016}.
Such decoupled approaches used pipeline models to implement MER and MEN separately, leading to errors cascade and absence of mutual benefits. 
There has been a line of research on joint modeling MER and MEN, which has demonstrated the superiority over pipeline implementation. 
For example, semi-CRF has been used for joint entity recognition and disambiguation \cite{Luo2015Joint}. \citeauthor{Leaman2016TaggerOne} (\citeyear{Leaman2016TaggerOne}) leverage a joint scoring function for MER and MEN. \citeauthor{Leaman2015tmChem} (\citeyear{Leaman2015tmChem}) developed a chemical named entity recognizer and normalizer created by combining two independent machine learning models in an ensemble. \citeauthor{Lou2017A} (\citeyear{Lou2017A}) propose a transition-based model to jointly perform disease named entity recognition and normalization. 

\textbf{Methodology of NER.}
Traditional approaches to NER include handcrafted features for Maximum Entropy models~\cite{Curran-Clark:2003:CONLL}, Conditional Random
Fields~\cite{McCallum-Li:2003:CONLL}, and Hidden Markov Models~\cite{Klein-etAl:2003:CONLL}.
State-of-the-art neural NER techniques use a combination of pre-trained word embeddings and character embeddings derived from a convolutional neural network (CNN) layer or bidirectional long short-term memory (Bi-LSTM) layer. These features are passed to a Bi-LSTM layer, which may be followed by a CRF layer~\cite{lample-EtAl:2016:N16-1,MaH16,TACL792}. \citeauthor{strubell-EtAl:2017:EMNLP2017} (\citeyear{strubell-EtAl:2017:EMNLP2017}) proposed a faster alternative
to Bi-LSTMs for NER: Iterated Dilated Convolutional Neural Networks (ID-CNNs), which have better capacity than traditional CNNs for large context and structured prediction.

\textbf{Neural Multi-Task Learning.} 
Multi-Task Learning  is a learning paradigm in machine learning and its aim is to leverage useful information
contained in multiple related tasks to help improve the generalization performance of all the tasks. It has been used successfully across many tasks of NLP \cite{Collobert2011}. 
In the context of deep learning for NLP, the most notable work was proposed by \citeauthor{CollobertW08} (\citeyear{CollobertW08}), which aims at solving multiple NLP tasks within one framework by sharing common word embeddings.
In recent years, the idea of neural deep multi-task learning becomes popular to sequence-to-sequence problems with LSTM~\cite{DongWHYW15,luong2016iclr_multi,Liuyang2016,AugensteinS17}. There are also a few studies which make use of multi-task learning for biomedical named entity recognition, such as cross-type biomedical named entity recognition~\cite{DBLP:journals/corr/abs-1801-09851} and multiple independent tasks modeling with MER involved~\cite{Crichton2017A}.

%% file: definition.tex
This section gives formal definitions of the two tasks to be investigated: MER and MEN.

\subsection{Medical Named Entity Recognition} 
The medical named entity recognition (MER) task is to find the boundaries of mentions from medical text. It differs from general NER in several ways. A large number of synonyms and alternate spellings of an entity cause explosion of word vocabulary sizes and reduce the efficiency of dictionary of medicine. Entities often consist of long sequences of tokens, making harder to detect boundaries exactly. It is very common to refer to entities also by abbreviations, sometimes non-standard and defined inside the text. Polysemy or ambiguity is pronounced: proteins (normally
class GENE) are also chemical components and depending on the context occasionally should be classified as class CHEMICAL; tokens that are sometimes of class SPECIES can be part of a longer entity of class DISEASE referring to the disease caused by the organism or the specialization of disease on the patient species.
In this work, we follow the setup of the shared subtasks of BioCreative \cite{Wei2015Overview}. Given a sentence $s$, i.e., a word sequence $w_1, ..., w_n$, each word is annotated with a predicated MER tag (e.g., ``B-DISEASE"). Therefore, we consider MER as a sequence-labeling task. 

\subsection{Medical Named Entity Normalization}
Medical named entity normalization (MEN) is to map obtained medical named entities into a controlled vocabulary. It
 is usually considered as a follow-up task of MER because MEN is usually conducted on the output of MER. In other words, MER and MEN are usually considered as hierarchical tasks in previous studies. In this paper, we consider MEN and MER as parallel tasks. MEN takes the same input with MER and have different output, i.e., for each word sequence $w_1, ..., w_n$, MEN outputs a sequence of tags from a different tag set. Therefore, we also consider MEN as a sequence-labeling task with the same input with MER. 
 
MER and MEN are not independent. MER and MEN are essentially hierarchical tasks but their outputs potentially have mutual enhancement effects for each other. Specifically, the output of MER, such as ``B-DISEASE", is a clear signal indicating the beginning of a disease entity, making MEN to map the code of a disease. Conversely,  the output of MEN, such as ``D054549-P" which is a disease code tag, is very helpful to recognize it as a part of a disease named entity.  

%% file: 3method.tex
Medical named entity recognition and normalization (MER and MEN) are hierarchical tasks and their outputs potentially have mutual benefits for each other as well. Specifically, the output of MER, such as ``B-DISEASE", is a clear signal indicating the beginning of a disease entity, leading to reducing the searching space of MEN and vice versa. Therefore, we propose to incorporate two explicit feedback strategies into multi-task learning framework to model mutual enhancement effects between tasks~\footnote{Our code is at GitHub (https://github.com/SendongZhao/Multi-Task-Learning-for-MER-and-MEN).}.
 In addition, we exploit Bi-LSTM to power the sequential modeling of the text and CNN to encode clues hidden in character-level features such as \textbf{Zo}lmitriptan, \textbf{Zomig} and \textbf{Zomig}on.

\textbf{Notation.} We use $\mathbf{x}_{1:n}$ to denote a sequence of $n$ vectors $\mathbf{x}_1,..., \mathbf{x}_n$. $F_{\theta}(\cdot)$ is a Bi-LSTM parameterized with parameters $\theta$. We use $F_L(\cdot)$ as a forward LSTM and $F_R(\cdot)$ as a backward LSTM with specific sets of parameters $\theta_L$ and $\theta_R$. $MER(w_{1:n},i)$ is the function to represent medical named entity recognition taking word sequence $w_{1:n}$ and index $i$ as input and output the corresponding named entity tag $y_{MER}^{i}$. $MEN(w_{1:n},i)$ is the function to represent medical named entity normalization taking word sequence $w_{1:n}$ and index $i$ as input and output the corresponding controlled vocabulary tag $y_{MEN}^{i}$.
We use $\circ$ to denote a vector concatenation operation. $\mathbf{U}$ and $\mathbf{V}$ are matrices to map the feedback of one task to the other. In this paper, we denote scalars by lowercase letters, such as $x$; vectors by boldface lowercase letters, such as $\mathbf{x}$; and matrices by boldface uppercase letters, such as $\mathbf{X}$.

\subsection{CNN for Character-level Representation}
\begin{figure}[h]
	\centering
	\includegraphics[width=0.45\textwidth]{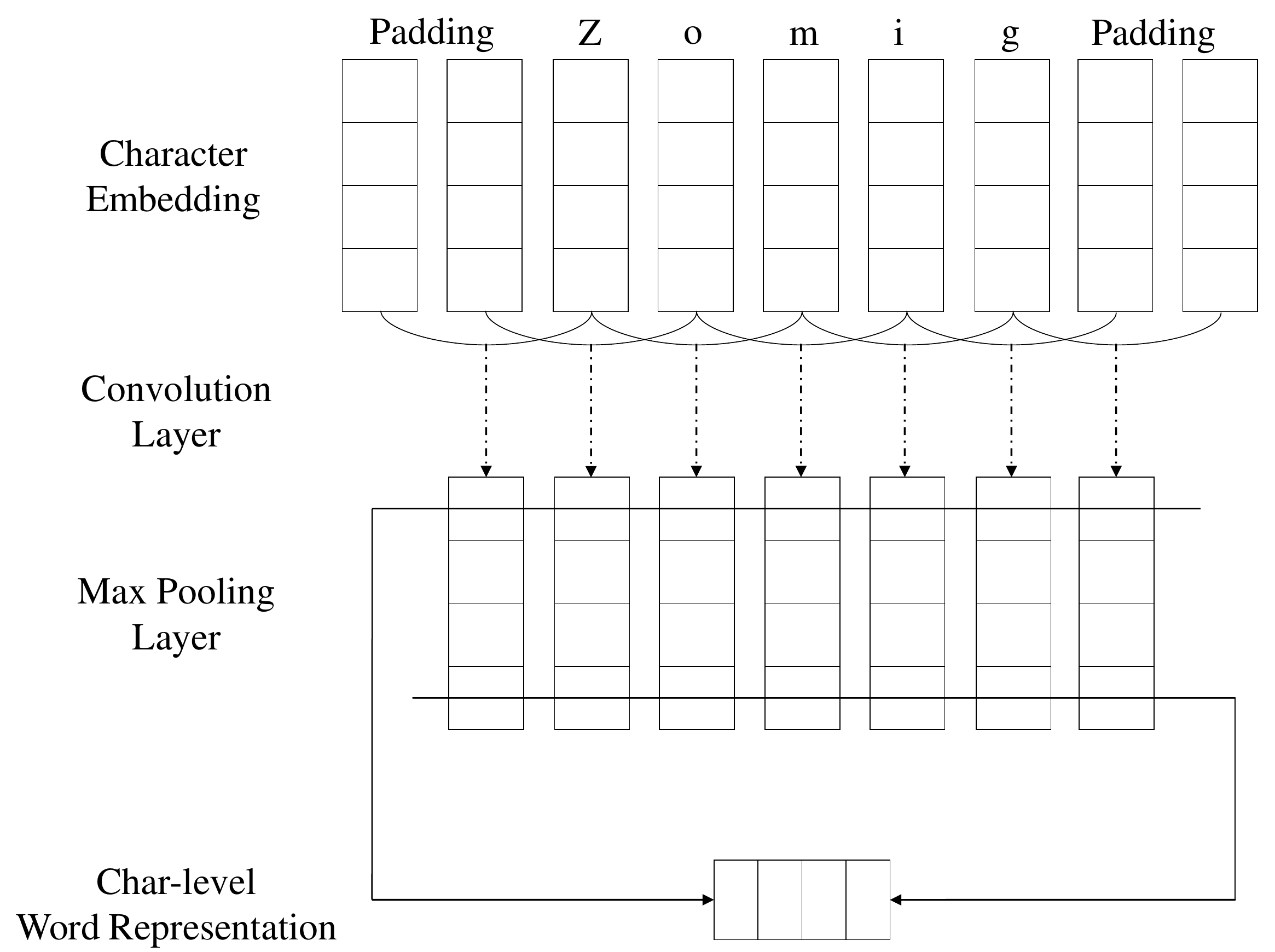}
	\vspace{-0.1in}	
	\caption{The CNN layer for extracting character-level word representation of word \textit{Zomig} (another name of the \textsc{Drug} \textit{Zolmitriptan} and \textit{Zomigon}).
		Dashed arrows indicate a dropout layer applied before character embeddings are fed into CNN. }\label{fig: cnn}
	\vspace{-0.15in}	
\end{figure}

Previous studies~\cite{TACL792} have shown that CNN is an effective approach to extract morphological
information (like the prefix or suffix of a word) from characters of words and encode it into neural representations. Figure~\ref{fig: cnn} shows the CNN we use to extract character-level representation of a given word. The CNN is similar to the one in~\citeauthor{TACL792} (\citeyear{TACL792}) except that we use only character embeddings as the inputs to CNN, without character type features. A dropout layer is applied before character embeddings are fed into CNN.

\subsection{Sequence-labeling with Bi-LSTM}

The extracted features of each word, including pre-trained word embeddings from Word2Vec and character-level word representation from CNN, are fed into a forward LSTM and a backward LSTM. The output of each network at each time step is decoded by a linear layer and a log-softmax layer into log-probabilities for each tag category. These two vectors are then simply added together to produce the final output.

\begin{figure*}[tp]
	\centering
	\includegraphics[width=0.95\textwidth]{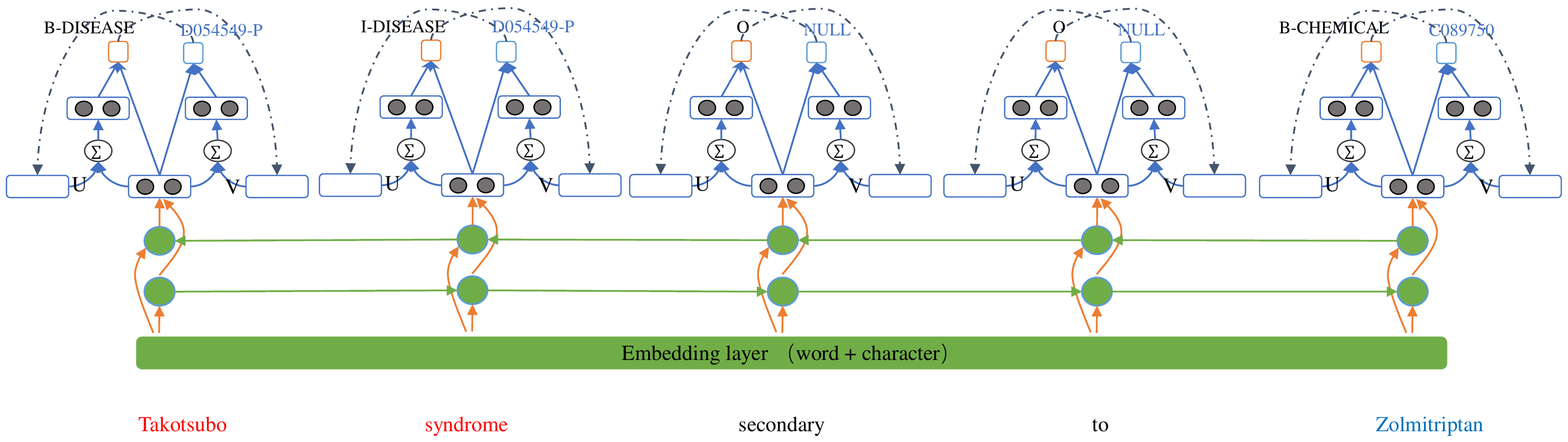}
	\vspace{-0.1in}	
	\caption{The main architecture of our neural
		multi-task learning model with two explicit feedback strategies for MER and MEN. The character embedding is computed by CNN in Figure~\ref{fig: cnn}.  Then
		the character representation vector is concatenated
		with the word embedding before feeding into the
		Bi-LSTM. Dashed arrows from the left to the  right is the feedback from MER to MEN.  Dashed arrows from the right to the left is the feedback from MEN to MER. Orange arrows indicate dropout
		layers applied on both the input and output vectors
		of Bi-LSTM. }\label{fig: model}
	\vspace{-0.15in}	
\end{figure*}

We view LSTM as a parameterized function $F_{\theta}(\mathbf{x}_{1:n})$ mapping a sequence of $n$ input vectors $\mathbf{x}_{1:n}$, $\mathbf{x}_i\in \mathbb{R}^{d_{in}} $ to output $n$ vectors $\mathbf{h}_{1:n}$, $\mathbf{h}_i \in \mathbb{R}^{d_{out}}$.
A Bi-LSTM is composed of two LSTMs, denoted as functions $F_L$ and $F_R$. One reading the sequence in its regular order, and the other reading it in reverse. Concretely, given a sequence $\mathbf{x}_{1:n}$ and
a desired index $i$, the function $F_{\theta}(\mathbf{x}_{1:n},i)$
is defined as:
\begin{equation*}
\begin{split}
F_{\theta}(\mathbf{x}_{1:n},i) = \mathbf{v}_i = \mathbf{h}_{L,i}\circ \mathbf{h}_{R,i}\\
\mathbf{h}_{L,i} = F_L(\mathbf{x}_1, \mathbf{x}_2, ..., \mathbf{x}_i)\\
\mathbf{h}_{R,i} = F_R(\mathbf{x}_n, \mathbf{x}_{n-1}, ..., \mathbf{x}_i)
\end{split}
\end{equation*}

The vector $\mathbf{v}_i = F_{\theta}(\mathbf{x}_{1:n},i)$ is then a representation of the $i$th item in $\mathbf{x}_{1:n}$, taking into account both the entire history $\mathbf{x}_{1:i}$ and the entire future $\mathbf{x}_{i:n}$.
Finally, in a deep Bi-LSTM, both $F_L$ and $F_R$ are k-layer LSTMs, and $F_{\theta}^{\ell}(\mathbf{x}_{1:n},i)= \mathbf{v}_i = \mathbf{h}_{L,i}^{\ell}\circ \mathbf{h}_{R,i}^{\ell}$.

In a sequence tagging task, we are given an input $w_1, ..., w_n$ and need to predict an output
$y_1, ..., y_n$, $y_i \in \mathbf{y}_{1:|L|}^i$, where $L$ is a label set of interest; i.e., in a medical named entity recognition task, $L$ is
the named entity tag set, and $y_i$ is the named entity tag for
word $w_i$ such as ``B-DISEASE".

If we take the inputs $\mathbf{x}_{1:n}$ to represent a sequence of sentence words $w_1, ..., w_n$, we can think of $\mathbf{v}_i = F_{\theta}(\mathbf{x}_{1:n}, i)$ as inducing an infinite window around a focus word $w_i$. We can then use $\mathbf{v}_i$ as an input to a multi-class classification function $f(\mathbf{v}_i)$, to assign a tag $\hat{y_i}$ to each input location $i$. The tagger is greedy: the tagging decisions are independent of each other. Alternatively, we can also feed the output vectors
of Bi-LSTM to the CRF layer to jointly decode the best tag sequence. Note that dropout layers are applied on both the input and output vectors of Bi-LSTM.


For a k-layer Bi-LSTM tagger for MER and MEN we get:
\begin{equation*}
\begin{split}
MER(w_{1:n},i)& = y_{MER}^{i}= \arg\max\mathbf{y}_{MER}^{i}\\
& = f_{MER}(\mathbf{v}_i^k)\\
MEN(w_{1:n},i) &=y_{MEN}^{i}= \arg\max\mathbf{y}_{MEN}^{i}\\
& = f_{MEN}(\mathbf{v}_i^k)\\
\end{split}
\end{equation*}
\vspace{-0.2in}
\begin{equation*}
\begin{split}
&\mathbf{v}_i^k = F_{\theta}^k(\mathbf{x}_{1:n},i)\\
&\mathbf{x}_{1:n} = E(w_1), E(w_2), ..., E(w_n)
\end{split}
\end{equation*}
where $E$ as an embedding function mapping each
word in the vocabulary into a d-dimensional
vector, $\mathbf{y}_{MER}^{i}$ is the log-probabilities vector with the length of MER tag space, $y_{MER}^{i}$ is the output tag of MER, $\mathbf{y}_{MEN}^{i}$ is the log-probabilities vector with the length of MEN tag space, $y_{MEN}^{i}$ is the output tag of MEN, 
and $\mathbf{v}_i^k$ is the output of the $k$th Bi-LSTM layer
as defined above. All the parameters are trained separately for MER and MEN because we model MER and MEN as different sequence labeling tasks.

\subsection{Multi-task Mode with Explicit Feedback Strategies}
The dependencies between MER and MEN inspire us to explore their potential mutual benefits.
In order to make the most of the mutual benefits between MER and MEN, we propose to feed the above mentioned Bi-LSTM and its variants into multi-task learning framework with two explicit feedback strategies, as shown in Figure~\ref{fig: model}. This method (1) is able to convert hierarchical tasks into parallel multi-task mode while maintaining mutual supports between tasks; (2) benefits from general representations of both tasks provided by multi-task learning; (3) is effective in determining boundaries of medical named entities through explicit feedback strategies thus improves the performance of both MER and MEN.

We experiment with a multi-task learning architecture based on stacked Bi-LSTM, CNNs and CRF. Multi-task learning can be seen as a way of regularizing model induction by sharing representations with other inductions. We use stacked Bi-LSTM-CNNs-CRF with task supervision from multiple tasks, sharing Bi-LSTM-CNNs layers among the tasks.

MER and MEN are hierarchical tasks and their outputs potentially have mutual benefits for each other as well. It means MEN can take MER results as input, while the results of MEN can be also useful for MER. However, MER and MEN can be implemented independently as different sequence tagging tasks.  
Therefore, we 1) follow the popular strategy of multi-task learning to share representations between MER and MEN; and 2) propose to use mutual feedback between MER and MEN, i.e., the result of MER is fed into the MEN as part of the input and the result of MEN is fed into the MER as part of the input. The multi-task learning with two explicit feedback strategies for MER and MEN is defined as:
\begin{equation*}
\begin{split}
MER(w_{1:n},i) &= y_{MER}^{i}=\arg\max\mathbf{y}_{MER}^{i}\\
&= f_{MER}(\mathbf{v}_i^{MER})\\
MEN(w_{1:n},i) &= y_{MEN}^{i}=\arg\max\mathbf{y}_{MEN}^{i}\\
& = f_{MEN}(\mathbf{v}_i^{MEN})\\
\end{split}
\end{equation*}
\vspace{-0.1in}
\begin{equation*}
\begin{split}
&\mathbf{v}_i^{MER} = \mathbf{v}_i^k \circ (\mathbf{v}_i^k+\mathbf{y}_{MEN}^{i}\mathbf{U})\\
&\mathbf{v}_i^{MEN} = \mathbf{v}_i^k \circ  (\mathbf{v}_i^k+\mathbf{y}_{MER}^{i}\mathbf{V})\\
&\mathbf{v}_i^k = F_{\theta}^k(\mathbf{x}_{1:n},i)\\
&\mathbf{x}_{1:n} = E(w_1), E(w_2), ..., E(w_n)
\end{split}
\end{equation*}
where $f_{MER}(\mathbf{v}_i^{MER})$ is the MER multi-class classification function and $f_{MEN}(\mathbf{v}_i^{MEN})$ the MEN multi-class classification function. $\mathbf{v}_i^{MER}$ is the input of MER multi-class classification function, which combines the output of the shared stacked Bi-LSTM-CNNs and the explicit feedback from MEN.  $\mathbf{v}_i^{MEN}$ is the input of MEN multi-class classification function, which combines the output of the shared Bi-LSTM-CNNs and the explicit feedback from MER. $\mathbf{U}$ is the matrix to map the feedback from MEN to MER, $\mathbf{V}$ maps the  feedback from MER to MEN. You can consider $(\mathbf{v}_i^k+\mathbf{y}_{MER}^{i}\mathbf{V})$ as a modification according to the feedback from MER, which could make $\mathbf{v}_i^k$ a better vector to get the correct label, the same as $\mathbf{v}_i^k+\mathbf{y}_{MEN}^{i}\mathbf{U}$.

In the multi-task learning setting, we have two prediction tasks over the same input vocabulary space. These two prediction tasks share k-layer Bi-LSTM-CNNs (i.e., hard parameter sharing). Each task has
its own output vocabulary (a task-specific tag set), but all of them map the length $n$ input sequence into a length $n$ output tag sequence.

\textbf{The Multi-task training protocol.} 
We assume to separate training set into $T$ different subsets corresponding to $T$ different tasks. We label $T$ different subsets as  $D_1, ..., D_T$, where
each $D_t$ contains pairs of input-output sequences ($w_{1:n}$, $y_{1:n}^t$), $w_i \in W$, $y_i^t \in L^t$. The input sets of words $W$ is shared across tasks, but the output sets (tag set) $L^t$ are task dependent.

At each step in the training process we choose a random task $t$, followed by a random training instance $(w_{1:n}, y_{1:n}^t)\in D_t$. We use the tagger of task $t$ to predict the labels $\hat{y}_i^t$, suffer a loss with respect to the true labels $y_i^t$ and update the output log-probabilities vector $\mathbf{y}_i^t$ of $w_{1:n}$ as well as the model parameters. 
If we choose MER at the very first step, we take the feedback from MEN as a log-probabilities vector with the initialization of each element having the same value $\frac{1}{|L^t|}$ and vice versa, where $|L^t|$ is the length of the tag set of task $t$.
Notice that a task $t$ (eg. MER and MEN in this paper) is associated with the stacked Bi-LSTM-CNNs. The update for a sample from task $t$ affects the parameters of $f_t$ and the shared $k$-layer functions $F_{\theta}^1, .., F_{\theta}^k$, but not the parameters of $f_{{t}'\neq t}$.
This asynchronous training protocol makes it possible to implement our model in distributed way.
We tried a synchronized way of training as well but did not lead to any difference in results.

%% file: 4experiment.tex
\subsection{Datasets} 
\begin{table*}[tb]
	\small
	\centering
	\begin{tabular}{l|c|ccc}
		\hline
		\multirow{2}{*}{Corpus} & \multirow{2}{*}{\# of Articles} & \multicolumn{3}{c}{Entity Types and Counts}\\
		\cline{3-5}
		\multirow{2}{*}{}&  \multirow{2}{*}{}& \# of Disease Mentions & \# of Chemical Mention& \# of Concepts\\
		\hline
		BC5CDR&1,500&12,852&15,935& 5,818\\
		NCBI &793&6,881&0& 1,049\\
		\hline
	\end{tabular}
	\vspace{-0.1in}
	\caption{Overall statistics of BC5CDR and the NCBI.}\label{tab: datasets}	
	\vspace{-0.1in}
\end{table*}
We evaluate the performance of the MTL models on two corpora: BC5CDR task corpus \cite{Li2016BioCreative} and the NCBI Disease corpus \cite{Rezarta2014NCBI}. The BC5CDR corpus contains 1500 PubMed abstracts, which are equally partitioned into three sections for training, development and test, respectively. A disease mention in each abstract is manually annotated with the concept identifier to which it refers to a controlled vocabulary. The NCBI Disease corpus consists of 793
PubMed abstracts, which are also separated into training (593), development (100) and test (100) subsets. The NCBI Disease corpus is annotated with disease mentions, using concept identifiers from either MeSH or OMIM. Table~\ref{tab: datasets} gives the statistics of the two corpora. Due to the limit of the vocabulary of chemical, we only consider mapping disease mentions to a controlled vocabulary of diseases.   
To map disease mentions to MeSH/OMIM concepts (IDs), we use the Comparative Toxicogenomics Database (CTD) MEDIC disease vocabulary, which consists of 9700 unique diseases described by more than 67 000 terms (including synonyms).

\subsection{Pre-trained word embeddings}
\label{sec: embeddings}
We initialized the word embedding matrix with four types of publicly available pre-trained word embeddings respectively. 
The first is Word2Vec 50-dimensional embeddings trained on the PubMed abstracts together with all the full-text articles from PubMed Central (PMC) \cite{Pyysalo2013b}. The second is GloVe 100-dimensional embeddings trained on 6 billion words from Wikipedia and web text~\cite{pennington-socher-manning}. The third  is Senna 50-dimensional embeddings trained on Wikipedia and Reuters RCV-1 corpus~\cite{Collobert2011}. The fourth is the randomly initialized 100-dimensional embeddings which are uniformly sampled from range $[-\sqrt{\frac{3}{dim}},+\sqrt{\frac{3}{dim}}]$, where $dim$ is the dimension of embeddings~\cite{He:2015:DDR:2919332.2919814}.

\subsection{Evaluation Metrics and Settings} 
We perform experiments for both medical named entity recognition and medical named entity normalization.
We utilize the evaluation kit\footnote{http://www.biocreative.org/tasks/biocreative-v/track-3-cdr} for evaluating model performances. 
Metrics measured are 
concept-level precision, recall and F1.
  
Our single- and multi-task networks are 3-layer, Bi-LSTM-CNNs with pre-trained word embeddings. For the neural multi-task learning model, we follow the training procedure outlined in Section~\ref{sec:method}. We use the word embeddings setup in Section~\ref{sec: embeddings}. Character embeddings are initialized with uniform samples from
$[-\sqrt{\frac{3}{dim}},+\sqrt{\frac{3}{dim}}]$, where we set $dim = 30$. 
 We follow \cite{S2014Deep} in using the same dimension for the hidden layers. We use a dropout rate of 0.5 and train these architectures with momentum SGD with the initial learning rate of 0.001 and momentum of 0.9 for 20 epochs.

\subsection{Main Results}
\label{sec: main}
The first part of Table~\ref{tab: main} illustrates the results of 5 previous top-performance systems for (medical) named entity recognition and normalization. 
Among these previous studies, LeadMine and IDCNN are pipeline models, while Dnorm, TaggerOne, and Transition-based Model are joint models.
From the first part, it is clear that the joint models perform better than the pipeline models on both corpora. 
The second part of the table presents comparisons of Bi-LSTM and its variants for MER and MEN. Adding CRF layer on both Bi-LSTM and Bi-LSTM-CNNs can not bring significant improvements. It might because the most of entity mentions in our data sets are single-word entities, i.e., entity with one word. CNN layer for char-level representation causes significant improvements on Bi-LSTM and its variants for both MER and MEN. 
Bi-LSTM-CNNs and Bi-LSTM-CNNs-CRF significantly outperform Bi-LSTM and Bi-LSTM-CRF respectively, showing that character-level word representations are important for both recognition and normalization. The improvents rely on two clues, 1) different medical entities usually have the same prefix and suffix, such as \textbf{acet}ate, \textbf{acet}one, antr\textbf{itis} and pharyng\textbf{itis}. Modeling such character-level information can benefit recognition; 2) different names which refer to the same medical entity usually share the same character fragments, such as \textbf{Zo}lmitriptan, \textbf{Zomig} and \textbf{Zomig}on. Modeling such character-level information can benefit normalization.
\begin{table*}[tb]
	\small
	\centering
	\begin{tabular}{l|cc|cc}
		\hline
		\multirow{2}{*}{Method} & \multicolumn{2}{c|}{ \textbf{NCBI}} & \multicolumn{2}{c}{\textbf{BC5CDR}}\\
		\multirow{2}{*}{}&  Recognition&Normalization&Recognition&Normalization\\
		\hline
		LeadMine \cite{Lowe2015LeadMineDI}&-&-&- &\textbf{0.8612}\\
		Dnorm \cite{Leaman2013DNorm} &0.7980& 0.7820 &- & 0.8064 \\
		TaggerOne \cite{Leaman2016TaggerOne} &\textbf{0.8290}& 0.8070&0.8260 & 0.8370\\
		Transition-based Model \cite{Lou2017A}            &0.8205& \textbf{0.8262}& \textbf{0.8382} & 0.8562 \\
		IDCNN \cite{strubell-EtAl:2017:EMNLP2017}  & 0.7983& 0.7425& 0.8011 & 0.8107 \\   
		\hline 
		Bi-LSTM         &0.8075&0.7934&0.8060&0.8136\\   
		Bi-LSTM-CRF     &0.8077&0.7933&0.8062&0.8136\\
		Bi-LSTM-CNNs    &0.8246&0.8059&0.8464&0.8447\\   
		Bi-LSTM-CNNs-CRF&0.8248&0.8061&0.8466&0.8449\\   
		\hline
		MTL                    && &  &   \\
		\hdashline
		+Bi-LSTM         &0.8532&0.8435&0.8321&0.8440\\
		+Bi-LSTM-CRF     &0.8532&0.8436&0.8321&0.8442\\   
		+Bi-LSTM-CNNs    &0.8647&0.8693&0.8632&0.8720\\   
		+Bi-LSTM-CNNs-CRF&0.8648&0.8693&0.8632&0.8722\\   
		\hline
		MTL-MEN\_feedback      && &  &  \\   
		\hdashline  
		+Bi-LSTM         &0.8574&0.8542&0.8419&0.8530  \\
		+Bi-LSTM-CRF     &0.8575&0.8542&0.8420&0.8532\\   
		+Bi-LSTM-CNNs    &0.8653&0.8706&0.8642&0.8813\\   
		+Bi-LSTM-CNNs-CRF&0.8654&0.8709&0.8645&0.8813\\   
		\hline
		MTL-MER\_feedback      && &   &   \\
		\hdashline
		+Bi-LSTM          &0.8637&0.8608&0.8474&0.8576 \\
		+Bi-LSTM-CRF      &0.8638&0.8609&0.8477&0.8576\\
		+Bi-LSTM-CNNs     &0.8723&0.8731&0.8736&0.8821\\
		+Bi-LSTM-CNNs-CRF &0.8725&0.8733&0.8739&0.8822\\
		\hline
		MTL-MEN\&MER\_feedback & &&    & \\
		\hdashline
		+Bi-LSTM          &0.8699&0.8657&0.8538&0.8645\\
		+Bi-LSTM-CRF      &0.8699&0.8658&0.8539&0.8647\\
		+Bi-LSTM-CNNs     &\textbf{0.8743}&\textbf{0.8823}&0.8762&\textbf{0.8917}\\
		+Bi-LSTM-CNNs-CRF &0.8743&0.8823 &\textbf{0.8763}&0.8917\\
		\hline
	\end{tabular}
	\vspace{-0.1in}	
	\caption{F1 score of medical named entity recognition and normalization on two corpora.}\label{tab: main}
	\vspace{-0.15in}
\end{table*}

From the third part of Table~\ref{tab: main}, we can see that MTL framework with Bi-LSTM and its variants significantly outperforms the pipeline use of Bi-LSTM and its variants, which indicates the contribution of general representations of MER and MEN provided by MTL.
The fourth part and fifth part of Table~\ref{tab: main} present the improvements by incorporating the feedback from MEN and the feedback from MER. Both feedback strategies can improve the performance of MER and MEN on both corpora, but the feedback from MER performs better. 
It makes sense because the original order of task hierarchy is from MER to MEN. Therefore, it is very natural that MEN needs more supports from MER than MER needs from MEN.
The last part of Table~\ref{tab: main} presents the results of Bi-LSTM and its variants in the MTL framework with both feedback of MEN and MER. This feedback-based MTL framework achieves the best result on each Bi-LSTM based models, indicating it is the best MTL framework on Bi-LSTM based models for MER and MEN.

\subsection{Effect of Dropout}
Table~\ref{tab: dropout} compares the results with and without dropout layers for training sets. All other
hyper-parameters and features remain the same as our best model in Table~\ref{tab: main}.
We observe slightly improvements for the two
tasks on both corpora. It confirms the function of dropout in reducing over-fitting reported by \citeauthor{Srivastava2014Dropout} (\citeyear{Srivastava2014Dropout}).
\begin{table}[h]
	\small
	\centering
	\begin{tabular}{l|c|c|c|c}
		\hline
		\multirow{2}{*}{}&\multicolumn{2}{c|}{\textbf{NCBI}}&\multicolumn{2}{c}{\textbf{BC5CDR}}\\
		\cline{2-5}
		\multirow{2}{*}{}& MER&MEN&MER&MEN\\
		\hline
		No&0.8669&0.8713&0.8722&0.8821\\
		Yes &\textbf{0.8743}&\textbf{0.8823}&\textbf{0.8763}&\textbf{0.8917}\\
		\hline
	\end{tabular}
	\vspace{-0.1in}
	\caption{Results with and without dropout on two
		tasks (F1 score for both MER and MEN).}\label{tab: dropout}	
	\vspace{-0.2in}
\end{table}

\subsection{Influence of Word Embeddings}
As mentioned in Section~\ref{sec: embeddings}, in order to test the importance of pre-trained word embeddings, we performed experiments with different sets of publicly published word embeddings, as well as a random sampling method, to initialize our model. Table~\ref{tab: embeddings} gives the performance of three different word embeddings, as well as the randomly sampled one.
According to the results in Table~\ref{tab: embeddings}, models using pre-trained word embeddings achieve a significant improvement as opposed to the ones using random embeddings. Both MER and MEN rely heavily on pre-trained embeddings. This is consistent with results of previous work~\cite{HuangXY15,TACL792}.

\begin{table*}[tb]
	\small
	\centering
	\begin{tabular}{l|c|c|c|c|c}
		\hline
		\multirow{2}{*}{\textbf{Embedding}}& \multirow{2}{*}{Dimension}&\multicolumn{2}{c|}{\textbf{NCBI}}&\multicolumn{2}{c}{\textbf{BC5CDR}}\\
		\cline{3-6}
		\multirow{2}{*}{}&  \multirow{2}{*}{}&Recognition&Normalization&Recognition&Normalization\\
		\hline
		Random&100&0.7532&0.7746&0.7665&0.7725\\
		Senna &50&0.7944&0.8016&0.7911&0.7966\\
		GloVe &100&0.7963&0.8042&0.8009&0.8062\\
		Word2Vec&50&\textbf{0.8743}&\textbf{0.8823}&\textbf{0.8763}&\textbf{0.8917}\\
		\hline
	\end{tabular}
	\vspace{-0.1in}
	\caption{Results with different choices of word embeddings on the two tasks (F1 score).}\label{tab: embeddings}	
	\vspace{-0.1in}
\end{table*}
For different pre-trained embeddings Word2Vec 50 dimensional embeddings achieve best results on both tasks.
 This is different from the results reported by \cite{MaH16}, where Glove achieved significantly better performance on NER than Word2Vec embedding.  Senna 50-dimensional embeddings obtain similar performance with Glove on MER and MEN, also significant behind Word2Vec. 
 One possible reason that Word2Vec is significantly better than the other two embeddings on MER and MEN is using domain related text for training embeddings. Word2Vec embeddings were trained on PubMed abstracts and the full-text articles. The other two embeddings which were trained on other domain text, leading to vocabulary mismatch of entities.

\subsection{Boundary Inconsistency Error Analysis}
Since we model MER and MEN as different sequence labeling tasks, the result of MER likely have different boundary with the result of MEN. Table~\ref{tab: boundary} compares the ratios of boundary inconsistency of MER and MEN on each test set of both corpora. It is clear that our proposed MTL framework with two feedback strategies on MER and MEN can significantly alleviate the boundary inconsistency of MER and MEN thus improve the performance.
\begin{table}[h]
	\small
	\centering
	\begin{tabularx}{0.47\textwidth}{X|c|c}
		\hline
		&\textbf{NCBI}&\textbf{BC5CDR}\\	
		\hline
		Bi-LSTM-CNNs-CRF&0.0635&0.0563\\
		\hdashline
		MTL &\multirow{2}{*}{0.0412}&\multirow{2}{*}{0.0383}\\
		+Bi-LSTM-CNNs-CRF&\multirow{2}{*}{}&\multirow{2}{*}{}\\
		\hdashline
		MTL-MEN\&MER\_feedback &\multirow{2}{*}{\textbf{0.0134}}&\multirow{2}{*}{\textbf{0.0114}}\\
		+Bi-LSTM-CNNs-CRF&\multirow{2}{*}{}&\multirow{2}{*}{}\\
		\hline
	\end{tabularx}
	\vspace{-0.1in}
	\caption{Ratios of the boundary inconsistency of MER and MEN on two test sets.}\label{tab: boundary}	
	\vspace{-0.2in}
\end{table}

\subsection{OOV Entities Error Analysis}
To better understand the behavior of our model, we perform error analysis on Out-of-Vocabulary words (OOV). Specifically, we partition each data set into four subsets — in-vocabulary words (IV), out-of-training-vocabulary words (OOTV), out-of-embedding-vocabulary words (OOEV) and
out-of-both-vocabulary words (OOBV). A word is considered IV if it appears in both the training
and embedding vocabulary, while OOBV if neither. OOTV words are the ones do not appear in
training set but in embedding vocabulary, while
OOEV are the ones do not appear in embedding vocabulary but in training set. 
An entity is considered as OOBV if there exists at lease
one word not in training set and at least one word
not in embedding vocabulary, and the other three
subsets can be done in similar manner. Table~\ref{tab: oov_stat} presents the statistics of the partition on each corpus.
The embedding we used is  pre-trained 50-dimensional embeddings in \cite{Pyysalo2013b}, the same as Section~\ref{sec: main}.
\begin{table}[h]
	\small
	\centering
	\begin{tabular}{l|c|c}
		\hline
		&\textbf{NCBI}&\textbf{BC5CDR}\\	
		\hline
		IV&987&5,421\\
		OOTV &33&127\\
		OOEV &16&33\\
		OOBV &10&142\\
		\hline
	\end{tabular}
	\vspace{-0.1in}
	\caption{Statistics of the partition on each test set.
		It lists the number of unique entities.}\label{tab: oov_stat}	
	\vspace{-0.2in}
\end{table}

Table~\ref{tab: oov_result} illustrates the performance of our best model
on different subsets of entities. 
The largest improvements appear on the IV and OOTV subsets of both the two corpora on both tasks. This demonstrates
that by feeding into multi-task learning framework with explicit feedback, our model
is more powerful on entities that appear in pre-trained embedding sets, which shows the superiority of our model to make better use of pre-trained word embeddings and deal with entities which do not appear in training set.
\begin{table}[h]
	\small
	\centering
	\begin{tabular}{l|c|c|c|c}
		\hline
		\multirow{2}{*}{}&\multicolumn{2}{c|}{\textbf{NCBI}}&\multicolumn{2}{c}{\textbf{BC5CDR}}\\
		\cline{2-5}
		\multirow{2}{*}{}& MER&MEN&MER&MEN\\
		\hline
		\multicolumn{5}{c}{Bi-LSTM-CNNs-CRF}\\
		\hdashline
		IV   &0.8451&0.8254&0.8738&0.8677\\
		OOTV &0.8046&0.8094&0.8279&0.8354\\
		OOEV &0.7776&0.8064&0.7835&0.7821\\
		OOBV &0.7221&0.7354&0.6937&0.7223\\
		\hline
		\multicolumn{5}{c}{MTL-MEN\&MER\_feedback+Bi-LSTM-CNNs-CRF}\\
		\hdashline
		IV   &0.8931&0.9017&0.9042&0.9136\\
		OOTV &0.8667&0.8753&0.8661&0.8832\\
		OOEV &0.8053&0.8132&0.8163&0.82217\\
		OOBV &0.7668&0.7713&0.7345&0.7804\\
		\hline	
	\end{tabular}
	\vspace{-0.1in}
	\caption{Comparison of performance of our model on different subsets of entities (F1 score).}\label{tab: oov_result}	
	\vspace{-0.2in}
\end{table}

%% file: 5conclusion.tex
We study the practical valuable task of MER and MEN.
They are fundamental tasks in medical literature mining because many developments in this area are related to these two tasks. 
Previous state-of-the-art studies have demonstrated that the mutual benefits between medical named entity recognition and normalization are very useful.
To make use of the mutual benefits in a more advanced and intelligent way, we proposed a novel deep neural multi-task learning framework with two explicit feedback strategies to jointly model MER and MEN. Our method can convert hierarchical tasks, i.e., MER and MEN, into parallel multi-task mode while maintaining mutual supports between tasks.
Experimental results indicate that our model outperforms previous state-of-the-art studies.

%% file: 6acknoledgments.tex
We thank anonymous reviewers for their insightful comments and suggestions. This work is supported by NSF IIS-1716432 and IIS-1750326.